\documentclass{article}
\usepackage{authblk}
\usepackage[english]{babel}

\usepackage[a4paper,top=2cm,bottom=2cm,left=3cm,right=3cm,marginparwidth=1.75cm]{geometry}

\usepackage{amsmath}
\usepackage{graphicx}
\usepackage[colorlinks=true, allcolors=blue]{hyperref}

\title{YASMIN: Yet Another State MachINe library for ROS 2}

\author[1]{Miguel Ángel González-Santamarta} 

\author[1]{Francisco Javier Rodríguez-Lera}

\author[1,2]{Camino Fernández Llamas}
\author[3]{Francisco Martín Rico}
\author[1,2]{Vicente Matellán Olivera}

\affil[1]{Robotics Group, Universidad de León, León (Spain)}

\affil[2]{RIS group - LAAS-CNRS, Toulouse, (France)}

\affil[3]{Intelligent Robotics Lab, Universidad Rey Juan Carlos, Madrid (Spain)}

\begin{document}
\maketitle

\begin{abstract} 
State machines are a common mechanism for defining behaviors in robots, defining them based on identifiable stages. There are several libraries available for easing the implementation of state machines in ROS 1, as SMACH or SMACC, but there are fewer alternatives for ROS 2. YASMIN is yet another library specifically designed for ROS 2 for easing the design of robotic behaviors using state machines. It is available in C++ and Python,  provides some default states to speed up the development, and a web viewer for monitoring the execution of the system and helping in the debugging.
\end{abstract}

\section{Proposal }

\begin{itemize}
    \item Title: YASMIN: Yet Another State MachINe for ROS 2
    \item Presenter: Miguel Á. González-Santamarta / Camino Fernández-Llamas
    \item Type: short (10 min)
    
\end{itemize}

\section{Description} 

A Decision-making system must effectively drive the behavior generation process by deciding which actions or commands need to be selectively activated or executed in autonomous robots. One widely extended approach is using finite state machines (FSM) for designing these systems. 

A state machine is a mathematical model for organizing a process based on transitions among states. State machines have been widely used in robotics to define autonomous robot behaviors understood as transitions from one state, where a set of actions is performed, to another state where a different set of actions or the parametrization of them is execution. These actions can be high-level robot tasks, such as navigation, text-to-speech or speech-to-text, or low-level actions as specific controllers. In the same way, state-based systems can be composed of several state machines that will change from one state to another executing the robot tasks. These systems may become really complex depending on the transitions between the states and the mechanism to select and execute them.

In ROS 1 there are popular libraries such as SMACH~\cite{bohren2010smach} or SMACC~\cite{smacc} for designing robotic behaviors based on FSMs. Initially, these libraries were not migrated to ROS 2, so we developed YASMIN (Yet Another State MachINe), an alternative to SMACH and SMACC, as well as a visual debugging tool to monitor their execution. The proposed presentation will relate to the lessons learned during the development process.

YASMIN is an open-source library for Python and C++ designed for ROS 2. Code can be found in a GitHub repository~\footnote{\url{https://github.com/uleroboticsgroup/yasmin}}. It is intended for designing robot behaviors using state machines. It is necessary to know the transitions and the states that compose a state machine. Besides, all states and nested state machines will share a blackboard, which is an easy-to-use storage employed to share data. 

There are several advantages to using YASMIN for designing robot actions for instance:
\begin{itemize}
    \item The library is fully integrated into ROS 2.
    \item It is available for Python and C++, the two languages used in ROS 2.
    \item Python version can be used for fast prototyping.
    \item There are some default states to improve the integration with ROS 2 and speed up the development.
    \item There is a uniform way to design tasks and integrate modules. State machines can represent high-level actions and states can describe minor tasks. Transitions represent the changes in action flow.
    \item All relevant data can be easily shared among all the states and nested state machines thanks to the use of blackboards.
    \item State machines can be canceled and stopped, which means stopping the current executing state.    
\end{itemize}

\subsection{web viewer}
\label{sec:viewer}

A web viewer is also distributed with YASMIN libraries, which allows monitoring of the execution of the state machines. This improves debugging from any device connected to the robot.

\begin{figure}
\centering
\includegraphics[width=1.0\textwidth]{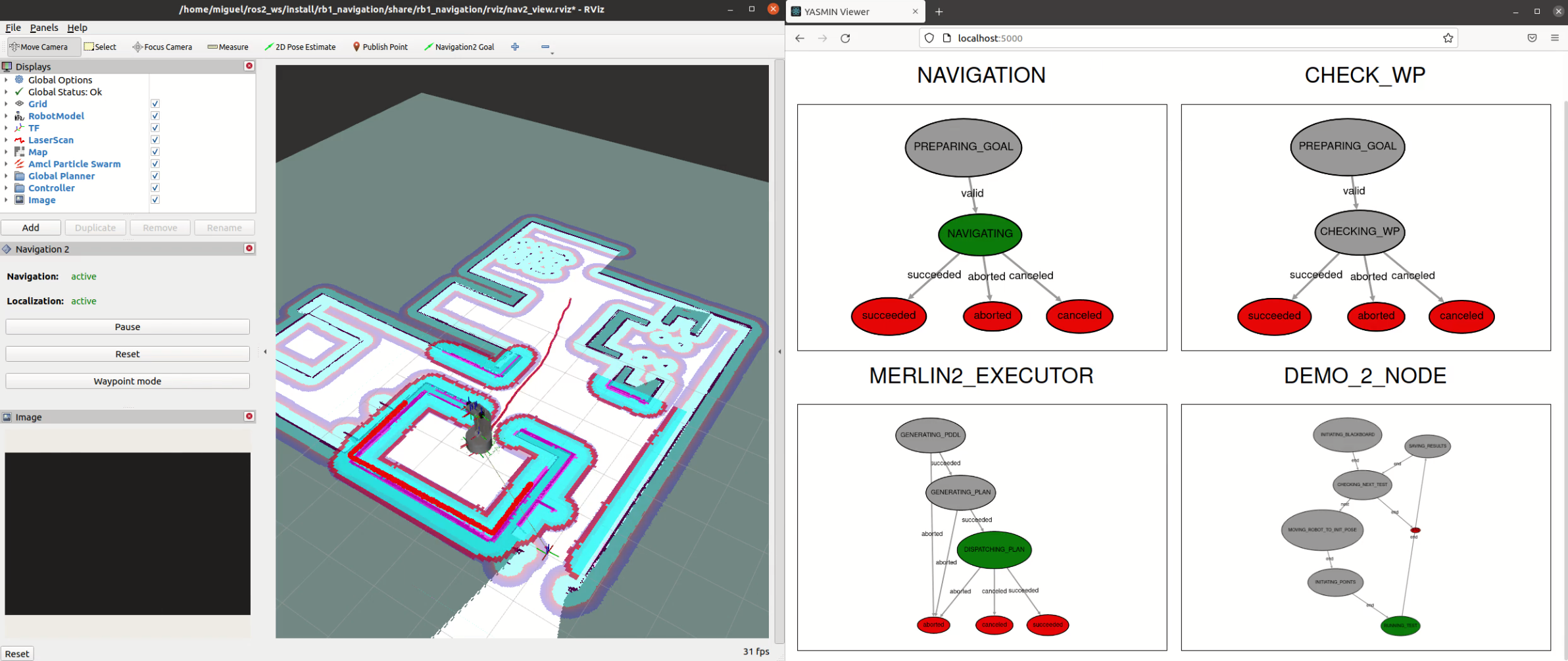}
\caption{\label{fig:viewer} YASMIN viewer.}
\end{figure}

Figure~\ref{fig:viewer} shows an example of a YASMIN web visualizer showing the state machines used in the robot shown on the left-hand side of the figure. In the shown example it uses two basic actions, and its corresponding state machines: navigating (NAVIGATION) and checking waypoints (CHECK\_WP). Besides, another state machine (MERLIN2\_EXECUTOR) is used to control the deliberative part of the control using a symbolic planner. Finally, the EMO\_2\_NODE state machine D is in charge of describing the goals of the robot at a high level. This means that its states may generate goals like "check waypoint one". 


\subsection{HFSM vs BTs}

YASMIN helps to implement robot behaviors using FSM. But, there are other mechanisms that can be used to implement behaviors in robotics. For instance, the use of behavior trees to produce modular structures is becoming very popular, and the use of a deliberative system based on symbolic planners has been a traditional approach.


Finite state machines (FSM) are a collection of states, where the execution of the actions in each state depends on the transitions happened. Hierarchical FSM (HFSM) let the generation of complex FSM, using hierarchy of choices. As a result, transitions can be organized better optimizing the code used.

In particular, the use of behavior trees (BT) has experienced a huge increase in robotics, \cite{colledanchise2021implementation}, \cite{iovino2022survey}. However, BTs are similar to HFSM except nodes are modular and uncoupled. Nodes can be executed from anywhere in the tree, allowing skipping irrelevant nodes. BTs can use a blackboard so that all nodes can share data.

FSMs, HFSMs and BTs are very similar but they should be used to treat different problems. BTs can produce more complex behavior structures thanks to their modular nature but can take more time to develop. In contrast, YASMIN allows designing FSMs and HFSMs that use blackboard easily and quickly. Finally, YASMIN is integrated into ROS 2 allowing the creation of HFSMs in Python and C++, the most commonly used languages for robotics.

\begin{figure}[ht!]
\centering
\includegraphics[width=0.8\textwidth]{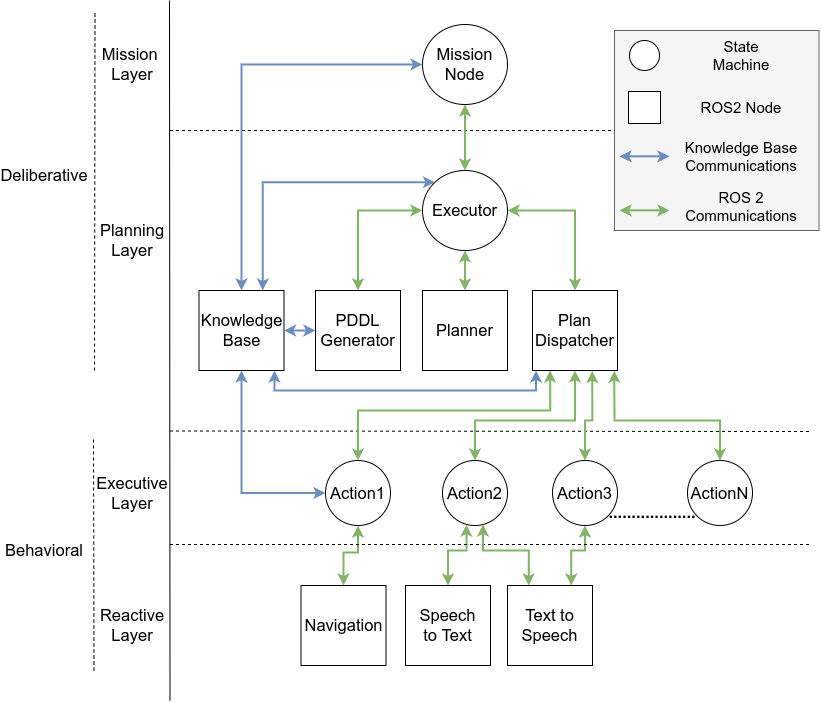}
\caption{\label{fig:merlin2} MERLIN2 Architecture.}
\end{figure}

\subsection{Integration in other systems}

YASMIN can be used as part of complex systems. In particular, it was originally implemented to be used in the cognitive architecture named MERLIN2 (MachinEd Ros 2 pLanINg), the new version of the  MERLIN~\cite{Gonzalez:2020} cognitive architecture. Figure~\ref{fig:merlin2} illustrates MERLIN2 components and shows how state machines defined using YASMIN are used in different layers:

\begin{itemize}
    \item Mission Layer: it manages the goals of the robot, which means creating goals that will be used by the Planning Layer to generate planes.

    \item Planing layer: it generates and executes plans. To do this, it generates the PDDL of the current state and domain. Afterward, that PDDL is used to generate the plan, the sequence of actions. Finally, the sequence of actions is executed by calling each action, which are the components of the Executive Layer.

    \item Executive Layer: is the layer composed of the actions used to produce plans. To improve the management of actions, they have been designed as state machines of YASMIN, whose states can process data or call systems of the Reactive Layer.

    \item  Reactive Layer: it is the layer composed of the reactive systems. This is the natural use of FSM, but it is not applied in this case. 
\end{itemize}

The example shown in section \ref{sec:viewer} corresponds to the use of MERLIN2 architecture in an @home robotic competition, showing the state machines of the executor, corresponding to the components in figure~\ref{fig:merlin2}.

\bibliographystyle{alpha}
\bibliography{sample}

\end{document}